\documentclass[10pt,twocolumn,letterpaper]{article}
\usepackage[pagenumbers]{include/cvpr/cvpr}
\usepackage{times}
\usepackage{epsfig}
\usepackage{graphicx}
\usepackage{amsmath}
\usepackage{amssymb}
\usepackage[breaklinks=true,bookmarks=false,colorlinks=true]{hyperref}
\usepackage{multirow}
\usepackage{inconsolata}
\usepackage{subcaption}
\usepackage{bbm}
\usepackage{algorithm}
\usepackage{algorithmicx}
\usepackage{algpseudocode}%
\usepackage{program}%
\usepackage{listings}%
\usepackage{float}
\usepackage{enumitem}
\usepackage{booktabs}
\usepackage{xcolor,colortbl}
\newcommand{\tsc}[1]{\textsuperscript{#1}}

\hypersetup{
    colorlinks=true,
    urlcolor={cyan},
    linkcolor={blue},
    citecolor={green}
}

\newcommand\extrafootertext[1]{%
    \bgroup
    \renewcommand\thefootnote{\fnsymbol{footnote}}%
    \renewcommand\thempfootnote{\fnsymbol{mpfootnote}}%
    \footnotetext[0]{#1}%
    \egroup
}

\title{Development and validation of an artificial intelligence model to accurately predict spinopelvic parameters}

\author{
Edward S. Harake\tsc{1}\quad
Joseph R. Linzey\tsc{1}\quad
Cheng Jiang\tsc{1}\quad
Rushikesh S. Joshi\tsc{1}\quad
Mark M. Zaki\tsc{1}\\
Jaes C. Jones\tsc{1}\quad
Siri S. Khalsa\tsc{2}\quad
John H. Lee\tsc{1}\quad
Zachary Wilseck\tsc{1}\quad
Jacob R. Joseph\tsc{1}\\
Todd C. Hollon\tsc{1}\quad
Paul Park\tsc{3}\\[1em]
\tsc{1}University of Michigan\quad
\tsc{2}Ohio State University\quad
\tsc{3}University of Tennessee\\[1em]
\url{https://mlins.org/spinepose/}
}

\begin{document}
\maketitle
\extrafootertext{Published as clinical article in \emph{Journal of Neurosurgery: Spine.} Correspondences to \href{mailto:tocho@med.umich.edu}{\texttt{tocho@med.umich.edu}}; \href{mailto:ppark@semmes-murphey.com}{\texttt{ppark@semmes-murphey.com}}.}

\begin{abstract}
\textbf{Objective.} Achieving appropriate spinopelvic alignment has been shown to be associated with improved clinical symptoms. However, measurement of spinopelvic radiographic parameters is time-intensive and interobserver reliability is a concern. Automated measurement tools have the promise of rapid and consistent measurements, but existing tools are still limited by some degree of manual user-entry requirements. This study presents a novel artificial intelligence (AI) tool called \textbf{SpinePose} that automatically predicts spinopelvic parameters with high accuracy without the need for manual entry.

\textbf{Methods.} SpinePose was trained and validated on 761 sagittal whole-spine X-rays to predict sagittal vertical axis (SVA), pelvic tilt (PT), pelvic incidence (PI), sacral slope (SS), lumbar lordosis (LL), T1-pelvic angle (T1PA), and L1-pelvic angle (L1PA). A separate test set of 40 X-rays was labeled by 4 reviewers, including fellowship-trained spine surgeons and a fellowship-trained radiologist with neuroradiology subspecialty certification. Median errors relative to the most senior reviewer were calculated to determine model accuracy on test images. Intraclass correlation coefficients (ICC) were used to assess inter-rater reliability.

\textbf{Results.} SpinePose exhibited the following median (interquartile range) parameter errors: SVA: 2.2 mm (2.3 mm), p = 0.93; PT: 1.3° (1.2°), p = 0.48; SS: 1.7° (2.2°), p = 0.64; PI: 2.2° (2.1°), p = 0.24; LL: 2.6° (4.0°), p = 0.89; T1PA: 1.1° (0.9°), p = 0.42; and L1PA: 1.4° (1.6°), p = 0.49. Model predictions also exhibited excellent reliability at all parameters (ICC: 0.91 - 1.0).

\textbf{Conclusions.} SpinePose accurately predicted spinopelvic parameters with excellent reliability comparable to fellowship-trained spine surgeons and neuroradiologists. Utilization of predictive AI tools in spinal imaging can substantially aid in patient selection and surgical planning. 


\textbf{Abbreviations.} AI = artificial intelligence; ASD = adult spinal deformity; CNN = convolutional neural networks; ICC = intraclass correlation coefficient; L1PA = L1 pelvic angle; LL = lumbar lordosis; PCK = percent correct keypoint; PI = pelvic incidence; PT = pelvic tilt; ROI = region of interest; SS = sacral slope; SVA= sagittal vertical axis; T1PA = T1 pelvic angle.
\end{abstract}
% ------------------------------------------------------------------------------
%----------------------------------------------------------------------------------------
\section{Introduction}
Adult spinal deformity (ASD) is a complex disease entity that can severely impact patients' lives in a myriad of ways. Spinopelvic parameters, generally derived from standing scoliosis radiographs, are an essential component of the diagnostic work-up, surgical treatment planning, and postoperative evaluation\cite{jackson1994radiographic, le2019sagittal, terran2013srs}. Appropriate spinal realignment in those with abnormal spinopelvic parameters can lead to significant improvements in quality of life\cite{liu2013validation, schwab2013radiographical,ha2016clinical,protopsaltis2020should}. Sagittal balance, in particular, is associated with improved quality of life for patients as measured by patient-reported outcomes\cite{glassman2005correlation, glassman2005impact, smith2013change, diebo2015sagittal, kieser2019validation}.A disadvantage with spinopelvic parameters is that they require manual measurements on standing scoliosis X-rays, which introduces variability to the evaluation process. While there is still some debate about the appropriate surgical approach merited for treating deformities, the ability to reliably and consistently measure spinopelvic parameters can improve the treatment of ASD patients.

Prior studies have attempted to tackle the problem of automated radiographic measurements with varying degrees of success. The earliest studies utilized surface topography with laser imaging combined with machine learning methodologies to derive information about scoliotic deformities and predict potential progression of scoliotic curves while avoiding excessive radiation exposure\cite{bergeron2005prediction,jaremko2001estimation,komeili2015monitoring,ramirez2006support,watanabe2019application,yang2019development}. Further studies used machine learning–based algorithms to begin estimating spinal anatomy and geometry from radiographs to assist with parameter predictions. More recently, artificial intelligence (AI) tools have been developed to assist in obtaining spinopelvic measurements. However, these early algorithms only utilized lateral lumbar X-rays \cite{schwartz2021deep,orosz2022novel}, limiting the number of spinopelvic parameters that could be measured. Additionally, some of these algorithms still are limited by some degree of manual user-entry requirements. This study presents a novel AI algorithm called SpinePose that automatically predicts spinopelvic parameters across multiple radiographic views without the need for any manual entry.

\begin{figure*}[h!]
    \centering
    \includegraphics[width=\textwidth]{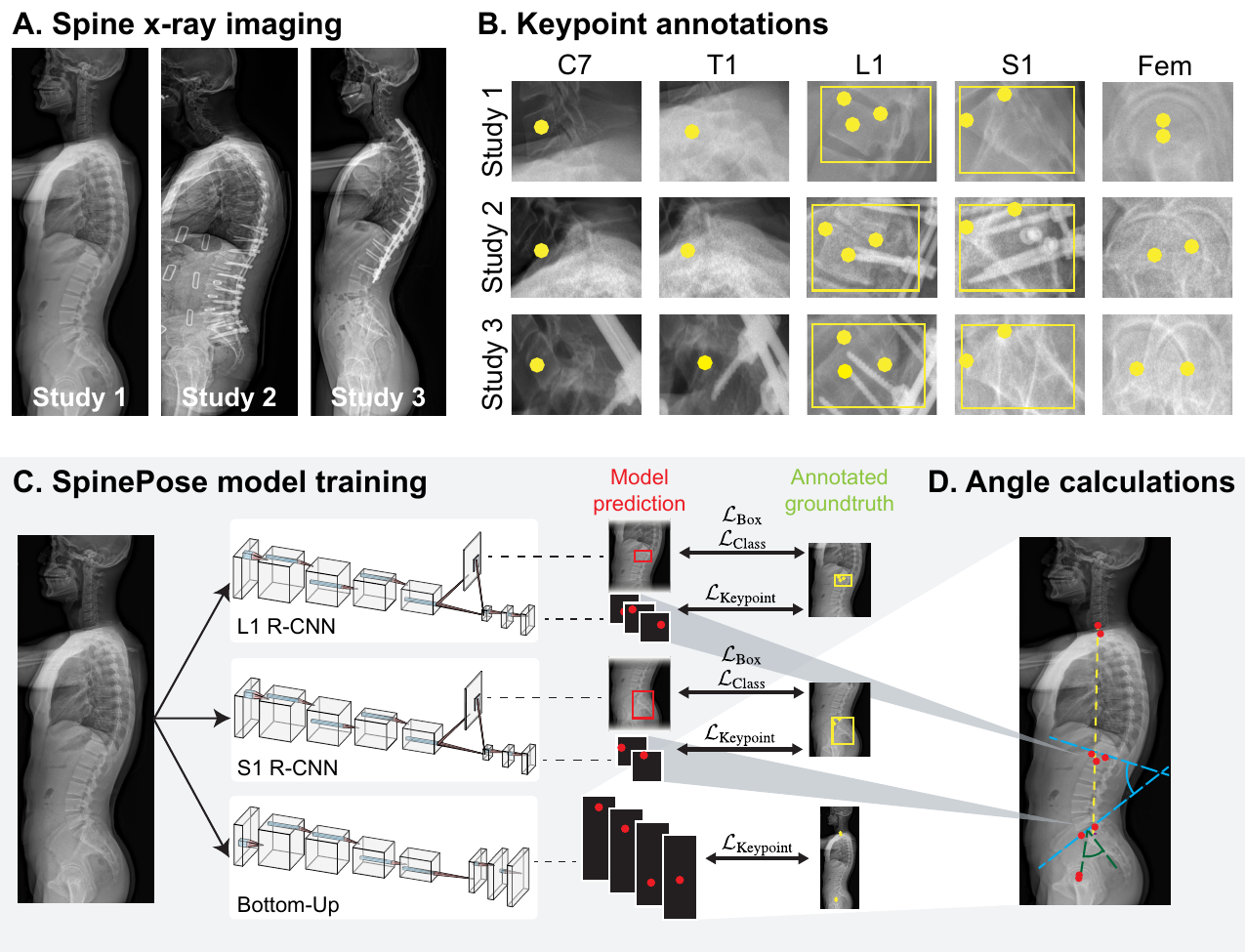}
    \caption{\textbf{SpinePose training pipeline.} (A) Standing whole-spine X-rays taken at a single academic institution were searched via an intra-institutional free-text search tool (EMERSE) and subsequently processed at the University of Michigan Radiology IT department. (B) Following image pre-processing, a senior Neurosurgery resident annotated each image with 9 total spinal keypoints at levels C7, T1, L1, S1, and both femoral heads. Bounding boxes were placed around regions L1 and S1. (C) Each input image was fed through 3 parallel convolutional neural networks: L1-model, S1-model, and R (remaining keypoints) model. The L1 and S1 models utilized a two-staged top-down approach facilitated by a region proposal network (RPN), whereas the R model used a bottom-up approach without need for the RPN. Regional detection and classification were all optimized by minimizing their respective losses. (D) The respective outputs of each of the 3 models were combined into 1 aggregate output, and spinopelvic parameters of interest were automatically calculated.}
    \label{fig:training}
\end{figure*}

\section{Methods}

\subsection{Dataset and Pre-processing}

This study was approved by our Institutional Review Board (HUM00147520). The aims of this study were 1) to train a deep learning model to automatically predict spinopelvic parameters with similar accuracy as fellowship-trained spine surgeons and neuroradiologists; 2) to show generalizable model performance to a heterogenous and broadly representative patient dataset, including those with instrumentation, varying ASD severity, and anatomic variances/transitional anatomy; and 3) to expand our model's performance to other clinically useful imaging modalities such as lumbosacral X-rays.  

Our dataset consisted of 761 sagittal whole-spine scoliosis X-rays taken in a cohort of patients from a single academic institution from 2018 to 2023. Images were included if they were standing sagittal whole-spine X-rays encapsulating vertebral levels C7, T1, L1, and S1 along with both femoral heads. Adequate visualization of a spinal level was defined as inclusion of the whole vertebral structure in the image without significant obstruction from instrumentation, image noise, or extreme brightness or contrast variations. Images were excluded for modalities other than whole-spine X-rays (e.g., lumbosacral films, CT, MRI), inadequate visualization or obstruction of structures of interest, and seated positioning of the patient. Additional radiographic information was collected, including the presence and general type of instrumentation (i.e., spinal hardware, brace, and hip arthroplasty) and the number of instrumented spinal levels. 

To obtain images, a HIPAA-compliant intra-institutional free-text search tool (EMERSE) was used to identify patients who underwent a whole-spine X-ray \cite{hanauer2015supporting}. After filtering based on inclusion and exclusion criteria, images were processed by the University of Michigan Radiology IT department and imported in DICOM format. We subsequently converted all the images to a TIFF file format for easy visualization and model training (Figure \ref{fig:training}). An appropriately trained neurosurgery resident labeled all of the de-identified images using Datatorch (Dundas, Ontario, Canada), an open-source annotation tool. Labels included a total of 9 landmark keypoints, with 3 keypoints at L1 (2 at the anterior and posterior borders of the superior endplate and 1 at the vertebral body midpoint), 2 keypoints at the S1 superior endplate, and 1 keypoint each at the midpoints of C7, T1, and the femoral heads. A bounding box was drawn around the L1 and S1 vertebral levels since these levels would be fed through their own neural networks for each region (Figure \ref{fig:training}). 

\subsection{Training and Validation}

\paragraph{Model Architecture.} SpinePose consists of 3 separately trained, parallel convolutional-neural networks (CNN). A CNN is a deep learning model specialized to identify hierarchical patterns and features in images at both local and global levels. These features are then used to perform a downstream task such as image classification or object detection. In SpinePose, there is a L1-specific model, S1-specific model, and R model (``remaining" keypoints), each of which predicts the keypoint locations at their respective anatomic regions (Figure \ref{fig:training}). Keypoints for levels L1 and S1 are given their own CNNs since these regions have more keypoints and less anatomic distinction compared to structures such as the round femoral heads. The L1 and S1 models utilize a ``top-down" region-based model, while the R model uses a ``bottom-up" encoder-decoder model to predict the remaining keypoints. A ``top-down" approach consists of two-steps in which the model first detects a box around the predicted location of a spinal level and then subsequently predicts keypoints within the bounds of that box. A ``bottom-up" approach, on the other hand, consists of a single step where the model predicts keypoints directly on the native image without prior object detection. Further technical details regarding model architecture and training can be found in the Online Appendix. 
	
\paragraph{Training Process.} Our dataset of images was randomly split with 92\% of the images used during training and 8\% held out for validation. Each input image to SpinePose was fed through the 3 parallel CNNs with its corresponding keypoint set (Figure \ref{fig:training}). During training, a keypoint loss value was calculated by comparing the predicted keypoints and the ground truth annotations (Figure \ref{fig:training}). Using stochastic gradient descent, SpinePose was optimized to iteratively reduce this loss value.  The final keypoint predictions from each of the 3 models were compiled at the end, and the following spinopelvic parameters were automatically calculated from the keypoints: sagittal vertical axis (SVA), pelvic tilt (PT), pelvic incidence (PI), sacral slope (SS), lumbar lordosis (LL), L1 pelvic angle (L1PA), and T1 pelvic angle (T1PA) (Figure \ref{fig:training}).

\paragraph{Testing Process.} A separate set of 40 whole-spine X-rays unseen during training or validation was used for testing. All images were imported with a pixel resolution of 3.730 pix/mm. These images were all labeled by 4 different reviewers: 2 fellowship-trained spine surgeons with 18 (R1) and 5 (R2) years of experience, respectively; a fellowship-trained neurointerventional radiologist (R3); and a senior neurosurgery resident (R4). Given significant professional experience, the annotations from R1 were used as the ground truth for the test set. To further inform the degree of heterogeneity in the dataset, clinical information was also recorded from each patient in the test set, including main clinical diagnosis, comorbidities that impact bone density (e.g., osteoporosis/penia), presence of instrumentation, number of instrumented spinal levels, and notable anatomic variants. 

While SpinePose included only standing whole-spine X-rays in the training set, a separate experiment was conducted to assess the model's ability to generalize to other imaging modalities such as lateral lumbosacral X-rays. To accomplish this, the model was trained on the same images; however, random image cropping was implemented in the training process to force the model to learn representations without a whole-spine view. This newly trained model was then evaluated on the 40-image test set with each image manually cropped from the thoracolumbar region to the sacro-femoral region. Given the smaller field of view, SpinePose was tasked with predicting only LL and SS in these lumbosacral images. 

\subsection{Data Analysis}

Keypoint detection accuracy in the model was assessed via the percent correct keypoint (PCK) metric. PCK is defined by the percentage of keypoint predictions whose distance from the ground truth falls within a given threshold (1-10 mm). To calculate the PCK, the raw pixel-level distances were calculated between predicted keypoint coordinates and ground truth coordinates. Given the original image resolution of 3.730 pixels/mm, the pixel distances were then converted to millimeters. The proportion of keypoint predictions within a given distance threshold was determined across all 40 test-set patients and at each spinal landmark. The accuracy of spinopelvic parameters was recorded as the mean and median differences between the model and ground truth at each parameter. A Wilcoxon rank sums test was performed at each parameter to assess the statistical significance of the errors. Additionally, an intraclass correlation coefficient (ICC) was subsequently found among the model, ground truth, and reviewers to assess the reliability of interrater scores at each parameter \cite{liljequist2019intraclass}. All programming for this project was written in Python using open-source tools from Pytorch, including the Keypoint R-CNN model. ICC was calculated using Excel (Microsoft, Redmond, WA). 

\section{Results}
\subsection{Training and Validation}

Assessing the general image characteristics of the training/validation set, 388 (50\%) images included spinal instrumentation, 55 (7.2\%) depicted a patient wearing a brace, and 63 (8.2\%) exhibited hardware from a hip arthroplasty/femoral fixation. From images with spinal screws/rods, the average (±SD) number of levels instrumented was 4.6 ± 3.4 levels (n = 368 images).

\begin{figure}[htb!]
    \centering
    \includegraphics[width=\columnwidth]{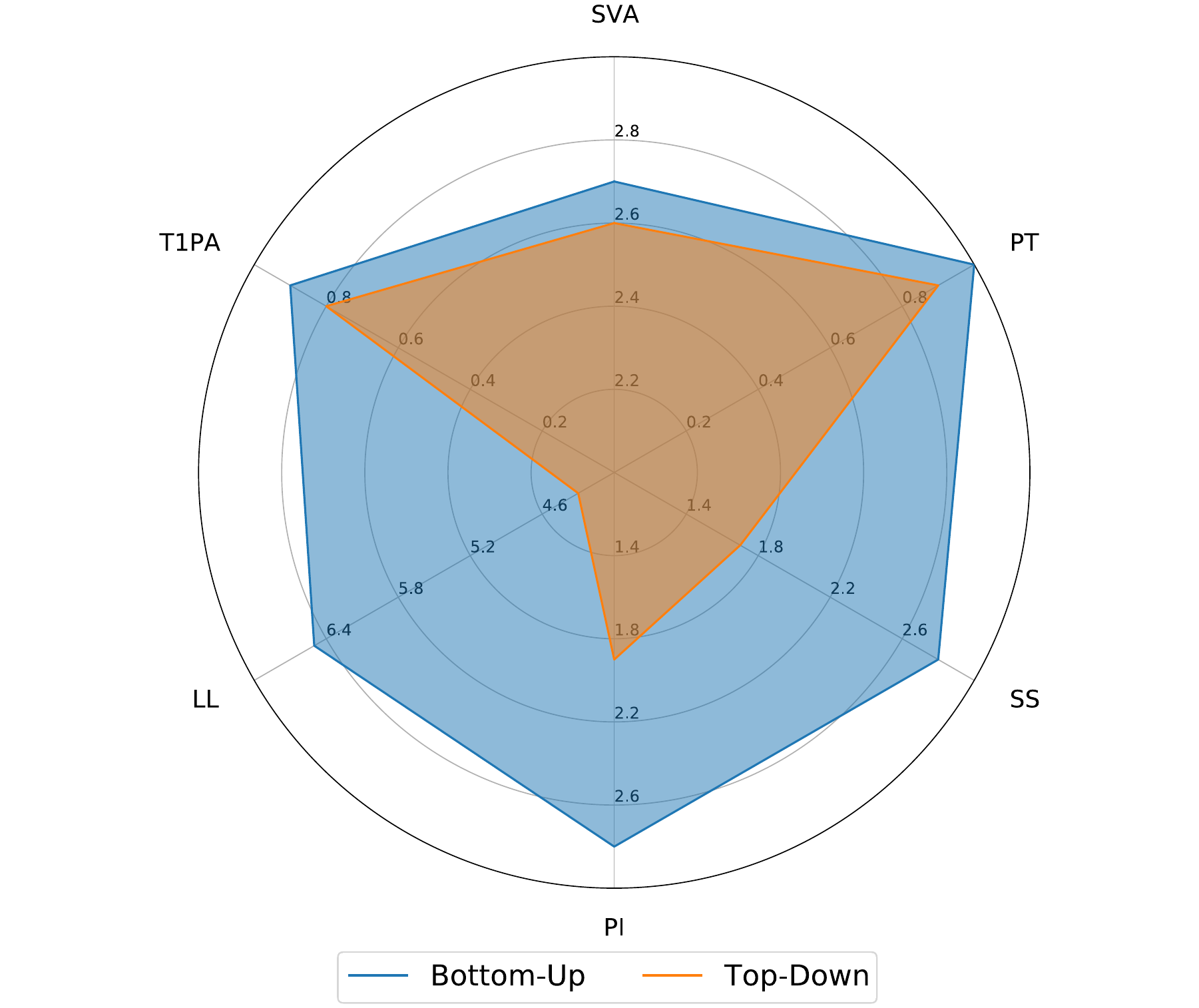}
    \caption{\textbf{Bottom-up vs. top-down training approach.} Radar plot comparing accuracy when using a single ``bottom-up" model vs. 3 parallel CNNs with a combined approach (SpinePose). The median errors of each training approach were compared relative to the ground truth. SpinePose exhibited lower median errors relative to ground truth annotations at all parameters. L1PA was not included in this analysis because the bottom-up model was not trained to predict this parameter. LL = lumbar lordosis; PI = pelvic incidence; PT = pelvic tilt; SS = sacral slope; SVA= sagittal vertical axis; T1PA = T1 pelvic angle.}
    \label{fig:radar}
\end{figure}

When using only a single ``bottom-up" model for all keypoint predictions, median parameter errors were as follows: SVA, 2.7 mm; PT, 1.0°; SS, 2.8°; PI, 2.8°; LL, 6.5°; and T1PA, 0.9° (Figure \ref{fig:radar}). These errors were higher at all spinopelvic parameters compared to using 3 parallel networks in SpinePose, which had the following median errors: SVA, 2.6 mm; PT, 0.9°; SS, 1.7°; PI, 1.9°; LL, 4.3°; and T1PA, 0.8° (Figure \ref{fig:radar}). 

\begin{table}[htb!]
    \centering
    \begin{tabular}{lc} \toprule
        Characteristics                                   & Patients\\\cmidrule(lr){1-1} \cmidrule(lr){2-2}
        \textbf{Total no.}                                & 40\\
        \textbf{Spinal pathologies}\\
           \hspace{1em}Degenerative disc disease          & 29 (72.5\%)\\
           \hspace{1em}Scoliosis                          & 23 (57.5\%)\\
           \hspace{1em}Stenosis                           & 18 (45.0\%)\\
           \hspace{1em}Stable spondylosis                 & 10 (25.0\%)\\
           \hspace{1em}Unstable spondylosis               & 5 (12.5\%)\\
        \textbf{Medical/surgical history}\\
           \hspace{1em}Osteoporosis/penia                 & 7 (17.5\%)\\
           \hspace{1em}Spine fusion                       & 22 (55.0\%)\\
           \hspace{1em}Other spine surgery                & 3 (7.5\%)\\
        \textbf{Instrumentation}\\
           \hspace{1em}Spine hardware                     & 23 (57.5\%)\\
           \hspace{1em}Brace                              & 5 (12.5\%)\\
           \hspace{1em}Hip arthroplasty/femoral fixation  & 5 (12.5\%)\\
        \textbf{Number of levels instrumented}            & 4.8 ± 4.1\\
        \textbf{Anatomic variants}\\
           \hspace{1em}Transitional vertebrae             & 3 (7.5\%)\\\bottomrule
    \end{tabular}
    \caption{Radiographic and clinical characteristics of test set patients.}
    \label{tab:testing_clinical}
\end{table}

\begin{table*}[ht!]
    \centering
    \begin{tabular}{ccccc}\toprule
        \multirow{2}{*}{Parameter} & \multicolumn{2}{c}{Ground Truth}
                                   & \multicolumn{2}{c}{Model-Predicted}\\\cmidrule(lr){2-3}\cmidrule(lr){4-5}
                    & Mean ( SD)     & Median (IQR)   & Mean ( SD)    & Median (IQR)\\\midrule
    SVA (mm)        & 24.1 (64.6)   & 16.8 (45.7)    & 23.8 (64.8)  & 16.7 (45.2)\\
    PT ($^\circ$)   & 14.9 (8.5)    & 13.5 (9.2)     & 15.9 (8.7)   & 14.5 (9.9)  \\
    SS ($^\circ$)   & 35.2 (10.9)   & 35.4 (14.8)    & 36.5 (11.8)  & 36.9 (16.3) \\
    PI ($^\circ$)   & 50.1 (10.0)   & 48.6 (12.6)    & 52.4 (10.3)  & 51.1 (10.5) \\
    LL ($^\circ$)   & 51.8 (17.5)   & 51.6 (26.4)    & 50.6 (17.1)  & 50.3 (26.4) \\
    T1PA ($^\circ$) & 12.1 (8.5)    & 10.8 (8.6)     & 13.1 (8.7)   & 12.4 (9.2)  \\
    L1PA ($^\circ$) & 6.9 (5.8)     & 6.2 (6.8)      & 7.7 (6.0)    & 6.3 (7.4)   \\\bottomrule
    \end{tabular}
    
    \caption{\textbf{Summary statistics of model vs. ground truth by parameter.} IQR, interquartile range; L1PA, L1 pelvic angle; LL, lumbar lordosis; PI, pelvic incidence; PT, pelvic tilt; SD, standard deviation; SS, sacral slope; SVA= sagittal vertical axis; T1PA, T1 pelvic angle.} \label{tab:summ_stats}
\end{table*}

\subsection{Testing}
Among all patients in the test set, the most common spinal pathologies included degenerative disc disease (73\%), scoliosis (58\%), and stenosis (45\%). Twenty-five (63\%) patients had a prior spine surgery history, with 23 (58\%) exhibiting spinal instrumentation on imaging (Table \ref{tab:testing_clinical}). In patients with spinal instrumentation, the average (±SD) number of instrumented levels was 4.8 $\pm$ 4.1 levels (Table \ref{tab:testing_clinical}). Five (12.5\%) patients also had hip/femoral instrumentation (Table \ref{tab:testing_clinical}). There were 3 (7.5\%) cases of transitional spinal variants including cervical and lumbosacral (Table \ref{tab:testing_clinical}). Using R1 as the ground truth, the average (±SD) values for sagittal balance in the test dataset were as follows: SVA, 24.1 mm ± 64.6 mm; PT, 14.9° ± 8.5°; SS, 35.2° ± 10.9°; PI, 50.1° ± 10.0°;  LL, 51.8° ± 17.5°; T1PA, 12.1° ± 8.5°; and L1PA, 6.9° ± 5.8° (Table \ref{tab:summ_stats}).

\begin{figure}[htb!]
    \centering
    \includegraphics[width=\columnwidth]{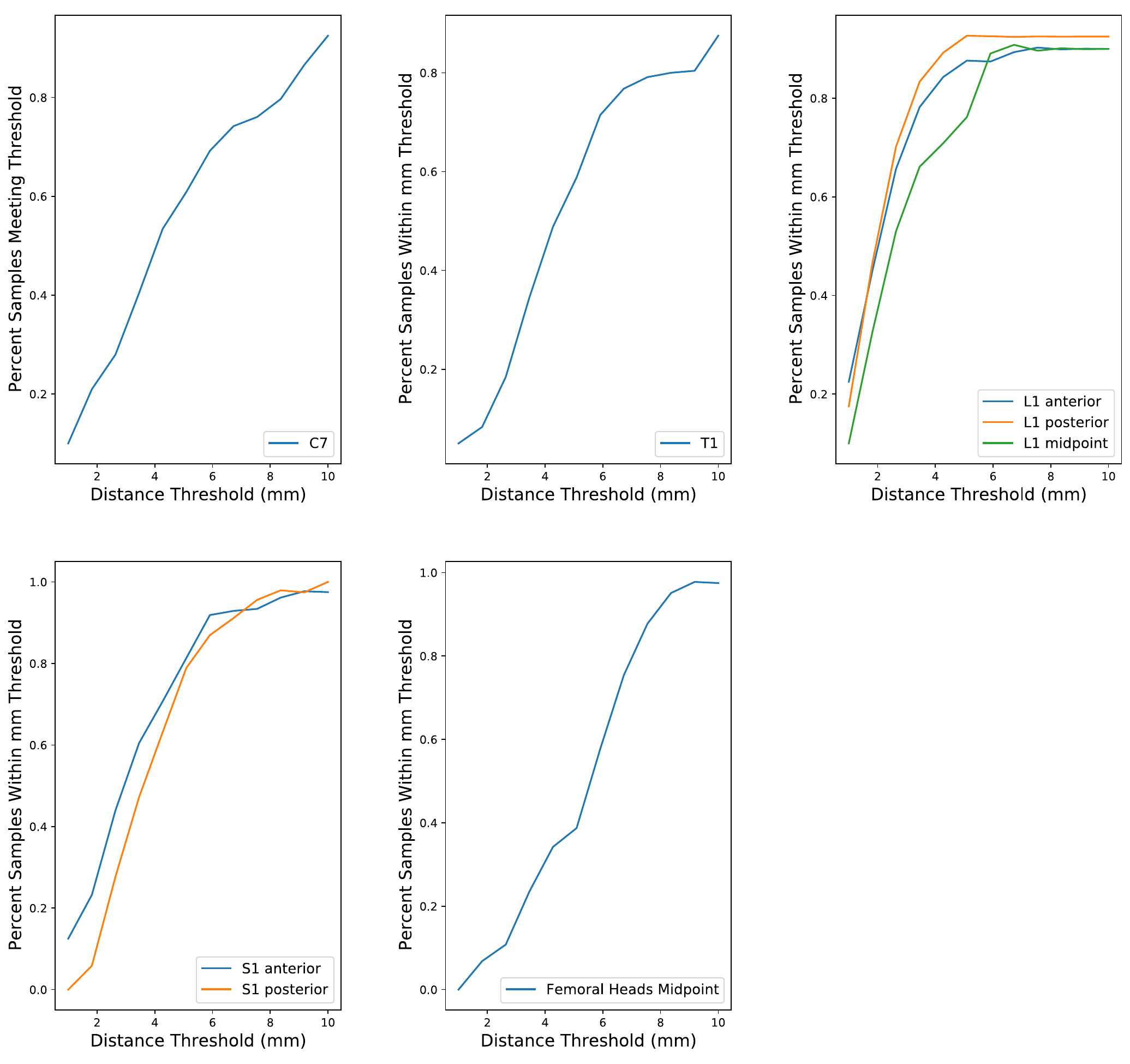}
    \caption{\textbf{Percent of keypoint predictions within different distance thresholds.} Plots depict keypoint detection accuracy across all spinal landmarks and a range of distance thresholds from ground truth (1-10 mm). *Note: The landmarks at each femoral head were combined into a single femoral midpoint landmark to account for difficulty in distinguishing right vs. left on a sagittal X-ray.}
    \label{fig:threshold}
\end{figure}

\definecolor{grey2}{HTML}{e9ecef}
\begin{table*}[ht!]
    \centering
    \begin{tabular}{cccccc} \toprule
        &\multicolumn{4}{c}{Model-Prediction Error} & Wilcoxon Rank Sums\\\cmidrule(lr){2-5}\cmidrule(lr){6-6}
        Parameter & \begin{tabular}{@{}c@{}}Overall Mean\\(SD)\end{tabular} &
        \begin{tabular}{@{}c@{}}Overall Median\\(IQR)\end{tabular} &
        \begin{tabular}{@{}c@{}}Median (IQR)\\With\\Instrumentation\end{tabular} &
        \begin{tabular}{@{}c@{}}Median (IQR)\\Without\\ Instrumentation\end{tabular} &
        p-value\\\midrule
        \multicolumn{6}{c}{\cellcolor{grey2}\textbf{Whole Spine Images}}\\
        SVA (mm) & 2.8 (2.9) & 2.2 (2.3) & 2.8 (1.7) & 1.3 (1.7) & 0.93\\
        PT (°)   & 1.4 (1.0) & 1.3 (1.2) & 1.2 (1.3) & 1.4 (1.0) & 0.48\\
        SS (°)   & 2.6 (3.6) & 1.7 (2.2) & 2.4 (1.7) & 0.8 (0.8) & 0.64\\
        PI (°)   & 3.0 (3.1) & 2.2 (2.1) & 1.9 (2.8) & 2.1 (1.2) & 0.24\\
        LL (°)   & 4.0 (4.5) & 2.6 (4.0) & 3.5 (6.8) & 1.8 (1.5) & 0.89\\
        T1PA (°) & 1.3 (0.8) & 1.1 (0.9) & 1.1 (0.8) & 1.2 (1.1) & 0.42\\
        L1PA (°) & 1.5 (1.2) & 1.4 (1.6) & 1.4 (1.5) & 1.3 (1.7) & 0.49\\\midrule
        \multicolumn{6}{c}{\cellcolor{grey2}\textbf{Lumbosacral Images}}\\
        SS (°) & 2.9 (4.6) & 1.9 (2.2) & 2.4 (1.6) & 1.1 (1.2) & 0.78\\
        LL (°) & 3.9 (4.8) & 2.9 (2.6) & 3.5 (2.1) & 1.5 (1.9) & 0.80\\\bottomrule
    \end{tabular}
    \caption{\textbf{Model error compared to ground truth based on image view and presence of instrumentation.} IQR = interquartile range; L1PA = L1 pelvic angle; LL = lumbar lordosis; PI = pelvic incidence; PT = pelvic tilt; SD = standard deviation; SS = sacral slope; SVA= sagittal vertical axis; T1PA = T1 pelvic angle.} \label{tab:instrumentation}
\end{table*}

\begin{figure}[htb!]
    \centering
    \includegraphics[width=\columnwidth]{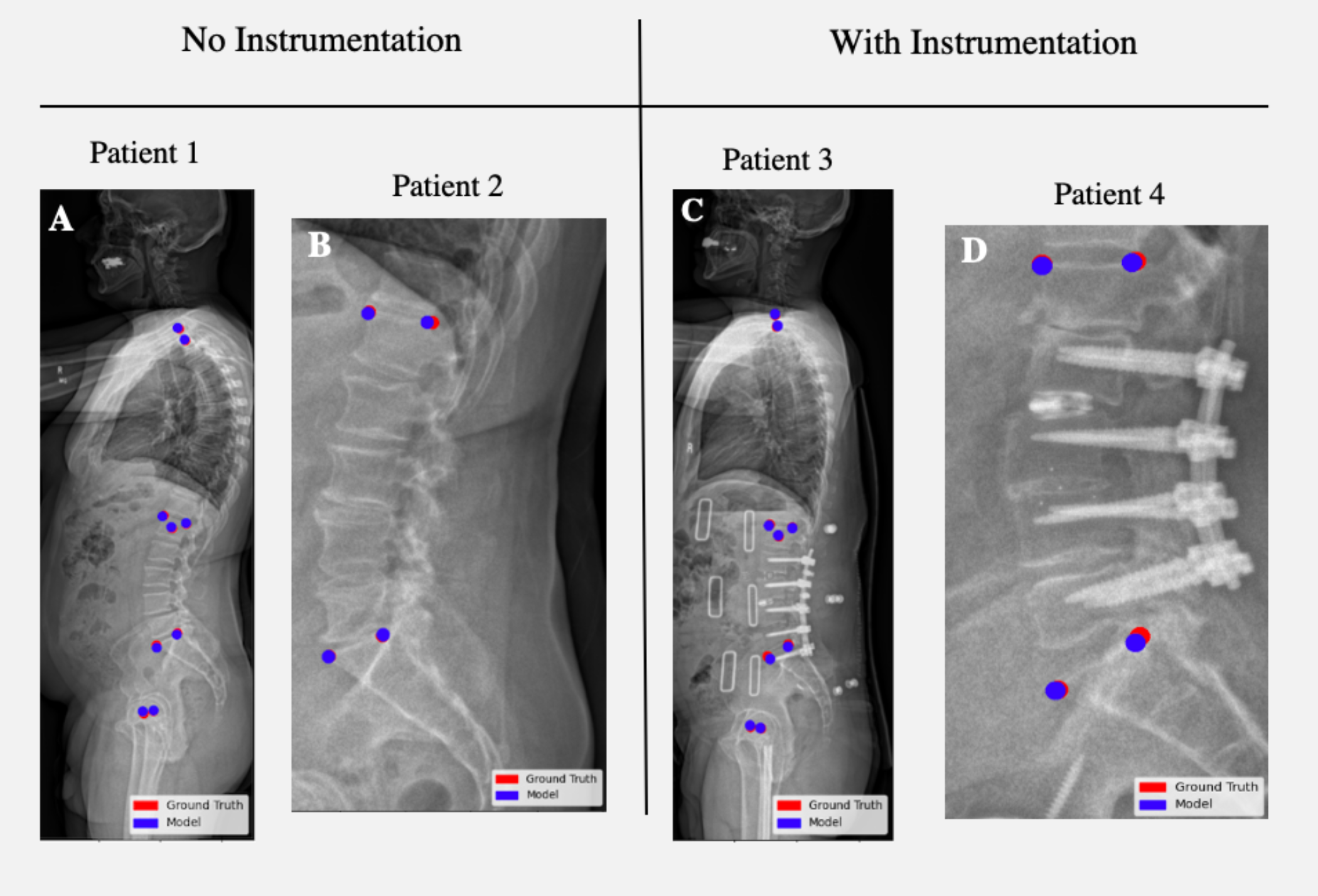}
    \caption{\textbf{Visualization of model vs. ground truth keypoint predictions in 4 patients.} Images A and B show a whole-spine and lumbosacral X-ray, respectively, without instrumentation. Images C and D show the same modalities with spinal instrumentation. }
    \label{fig:pred_vis}
\end{figure}

\begin{figure}[htb!]
    \centering
    \includegraphics[width=\columnwidth]{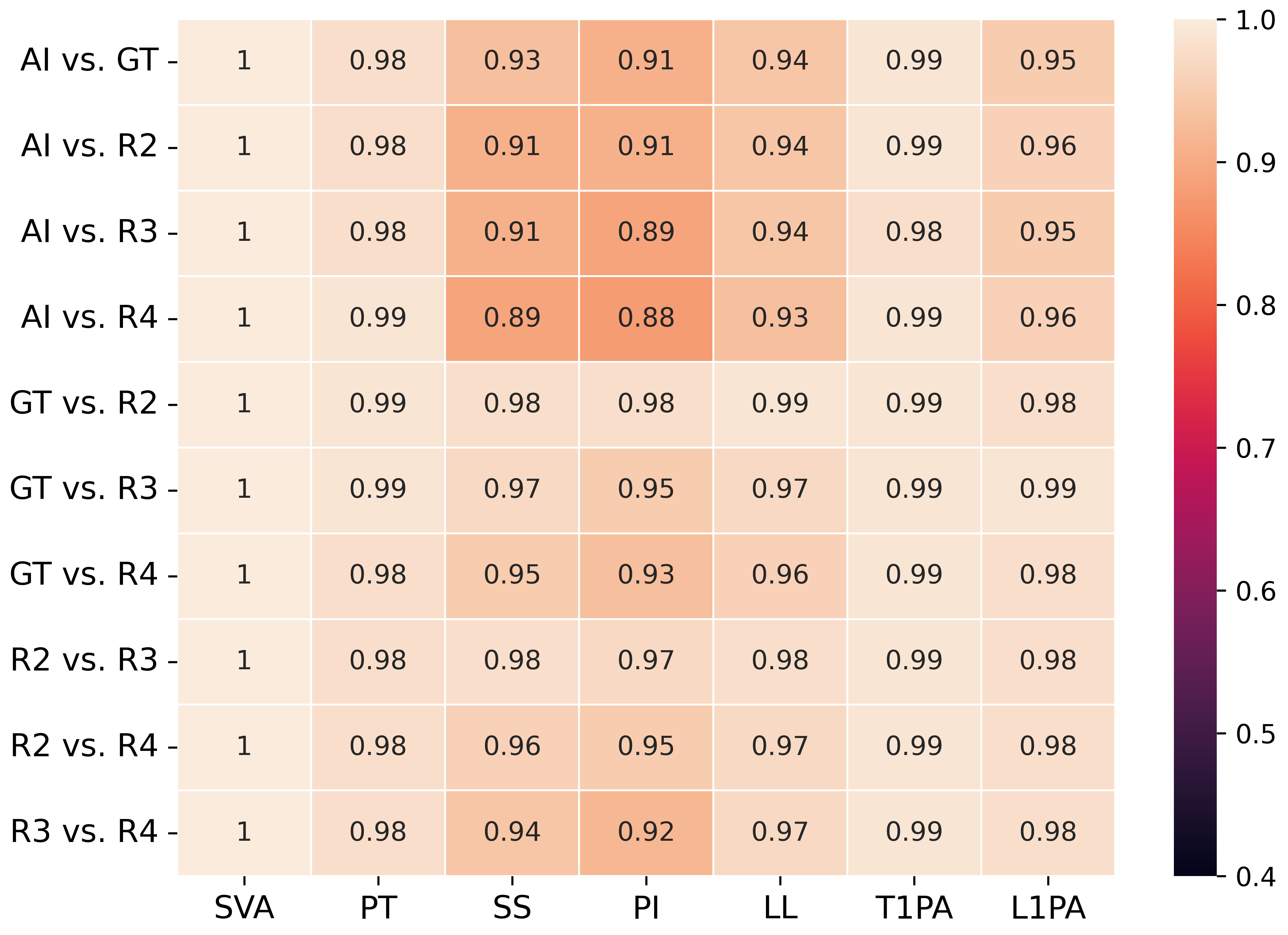}
    \caption{\textbf{Intraclass correlation coefficient (ICC) heatmap.} An ICC was calculated at each parameter between 2 separate raters. On a scale of 0-1, the ICC reflects inter-rater similarity among scores within a given class. SpinePose (AI) shows excellent reliability at all parameters when compared to ground truth (GT) as well as to each of the 3 remaining raters (R2-R4), who were a fellowship-trained spine surgeon (R2), a neuroradiologist (R3), and a senior neurosurgery resident (R4). L1PA = L1 pelvic angle; LL = lumbar lordosis; PI = pelvic incidence; PT = pelvic tilt; SS = sacral slope; SVA= sagittal vertical axis; T1PA = T1 pelvic angle.}
    \label{fig:icc}
\end{figure}

Analyzing keypoint prediction accuracy, SpinePose achieved high PCK values at a 5-mm threshold with greater accuracy at the L1/S1 landmarks (75-93\%) vs. the remaining landmarks (38-60\%) (Figure \ref{fig:threshold}). At the 10-mm threshold, nearly all keypoint predictions were successful (PCK 88-100\%) (Figure \ref{fig:threshold}). Looking at spinopelvic parameters, model predictions were similar to ground truth with no statistically significant difference (Table \ref{tab:instrumentation}; Figure \ref{fig:pred_vis}). Overall median (interquartile range) errors of SpinePose were: SVA, 2.2 mm (2.3 mm); PT, 1.3° (1.2°); SS, 1.7° (2.2°); PI, 2.2° (2.1°); LL, 2.6° (4.0°); T1PA, 1.1° (0.9°); and L1PA, 1.4° (1.6°). These predictions all showed excellent reliability compared to the ground truth readings (ICC: 0.91-1.0) and the other 3 reviewers (ICC: 0.88-1.0) (Figure \ref{fig:icc}). 

Looking specifically at patients with spinal instrumentation, median errors were: SVA, 2.8 mm (1.7 mm); PT, 1.2° (1.3°); SS, 2.4° (1.7°); PI, 1.9° (2.8°); LL, 3.5° (6.8°); T1PA, 1.1° (0.8°); and L1PA, 1.4° (1.5°). Model performance was also maintained in patients with transitional vertebral anatomy (Table \ref{tab:testing_clinical}), with median errors of SVA, 0.1 mm (1.5 mm); PT, 1.0° (1.7°); SS, 1.6° (7.8°); PI, 2.2° (5.5°); LL, 1.6° (8.1°), T1PA, 0.7° (1.4°), and L1PA, 0.7° (1.9°). 

SpinePose exhibited similarly high accuracy on lumbosacral X-rays with median errors: LL, 2.9° (2.6°), p = 0.80; SS, 1.9° (2.2°), p = 0.78 (Table \ref{tab:instrumentation}; Figure \ref{fig:pred_vis}). Accuracy remained high with (n = 22) and without (n = 18) instrumentation in the image (Table \ref{tab:instrumentation}). Reliability remained excellent compared to the ground truth (ICC: 0.92-0.93).

\section{Discussion}

While our understanding of ASD as a surgical disease continues to evolve, there is a clear need for standardization of the evaluation and treatment of this nuanced and highly variable disease process. Part of the variability in evaluating ASD is the radiographic measurement of spinopelvic parameters, particularly measurements of sagittal imbalance, which have repeatedly been shown to correlate highly to patients' quality of life and functional status\cite{jackson1994radiographic, le2019sagittal, terran2013srs, liu2013validation, schwab2013radiographical, ha2016clinical, protopsaltis2020should}. One challenge with treatment of ASD is the lack of a tool for uniform and consistent radiographic measurement of spinopelvic parameters. Presently, the standard is manual measurements that depend on user expertise, and as such are subject to user error and inter-user variability. In an effort to streamline and standardize this process, our group has developed a novel artificial intelligence–based algorithm called SpinePose to automate characterization of spinopelvic parameters.
 
Prior groups have attempted to utilize different algorithms to automate spinopelvic parameter measurement with varying degrees of success. Orosz et al. \cite{orosz2022novel} and Schwartz et al. \cite{schwartz2021deep} demonstrated highly accurate models to measure spinopelvic parameters. However, both of these models were limited to only lateral lumbosacral radiographs, and as such their measurements were restricted to LL, PI, PT, and SS. These algorithms have not been tested on whole-spine scoliosis films, which limits their ability to appropriately assess ASD. In particular, the omission of whole-spine radiographs prevents both studies from assessing measurements of global spinal alignment, including SVA and T1PA, which are instrumental in the evaluation of spinal deformity patients. Additionally, of the images utilized by Schwartz et al. \cite{schwartz2021deep}, $<$10\% had implants or instrumentation, whereas $>$50\% of our test set had instrumentation. Similarly, in the existing literature, Galbusera et al. \cite{galbusera2019fully} attempted to predict spinopelvic parameters (Cobb angle, SS, PI, PT, LL, and thoracic kyphosis) using biplanar radiographs, however this required performing 3D reconstructions of the spine from the radiographs, as well as image alterations for resizing images and altering aspect ratios throughout their analysis to prevent impaired parameter prediction. These steps introduce additional variables and/or bias to predictions. Their study also utilized limited images with instrumentation, and excluded transitional anatomy, which coupled with the omission of SVA and other parameters, further limits usefulness. In another study, Weng et al. \cite{weng2019artificial} demonstrated an excellent study with a large training/validation sample size (n = 990), a large proportion of images with instrumentation, and inclusion of patients with anatomical variance. However, their algorithm was limited to measuring only SVA with a high degree of accuracy, omitting all other spinopelvic parameters.
 
The advantage of our algorithm lies in the ability to predict spinopelvic parameters from both whole-spine lateral scoliosis radiographs and upright lumbosacral radiographs. A key novelty of SpinePose was its focus on only 9 core anatomic landmark keypoints. Including a high representation of anatomically heterogeneous images in the training set allowed optimization of model accuracy while preserving architectural simplicity. 

A recent study by Yeh et al. \cite{yeh2021deep} similarly sought to predict spinopelvic parameters using whole-spine X-rays. Their model required over 2000 images for training, and a far more computationally intensive and complex algorithm that estimated 45 anatomic landmarks to predict 18 different parameters, compared to the smaller dataset and 9 anatomic landmarks utilized in our study. Despite using a fraction of the data and model complexity employed by Yeh et al. \cite{yeh2021deep}, our study appears to have achieved similar or better accuracy with key parameters (SVA, PT, SS, PI, LL, T1PA), while importantly demonstrating better reliability when assessing ICC compared to the ground truth. All of our algorithm's predicted measurements achieved higher ICC values and fell in the range of `excellent' interobserver reliability (0.9 - 1.0), while three important parameters (SS, PI, and LL) fell out of this category in their study.

Our final training model combined 3 parallel CNNs consisting of a L1 and S1 model (using a ``top-down" region-based approach) and a R model (using a ``bottom-up" approach) to generate anatomic keypoints for downstream parameter prediction. The combination of these models yielded higher accuracy across all spinopelvic parameters compared to using 1 model with an entirely ``bottom-up" approach (Figures 1 and 2). Comparison of predicted parameter values to our ground truth demonstrates a high degree of accuracy in images both with and without spinal instrumentation (Table \ref{tab:instrumentation}) as well as transitional vertebral anatomy (Table \ref{tab:testing_clinical}). In addition, in the test set of 40 scoliosis lateral X-rays, $>$50\% of images had spinal instrumentation (Table \ref{tab:testing_clinical}), and our algorithm still achieved excellent predictive accuracy and reliability relative to fellowship-trained spine surgeons and neuroradiologists (Table \ref{tab:instrumentation}; Figure \ref{fig:pred_vis}). 

There are several limitations in both our study and similar prior studies that need continued exploration in future work. A host of spinopelvic parameters that are used for both research and clinical purposes were not explored in this initial study, including cervical SVA, C2 - C7 lordosis, thoracic kyphosis, and T1 slope. In addition, our algorithm needs external validation with testing on images from other institutions. 

\section{Conclusion}
This study presents a novel artificial intelligence algorithm to automatically identify key spinopelvic parameters, including lumbar lordosis, pelvic tilt, pelvic incidence, sacral slope, sagittal vertical axis, T1 pelvic angle, and L1 pelvic angle, from standing sagittal scoliosis X-rays. We believe the implementation of such processes can help standardize clinical assessment of patients with ASD and also assist with research efforts that require analysis of radiographic measurements for spine patients.

\section*{Acknowledgements and Conflicts-of-Interest}
We would like to thank Jim Good at the University of Michigan Radiology IT department for his invaluable help in image collection.

This work is supported in part by grants NIH T32TR004371, and NIH T32GM141746.

There are no conflicts of interest to report from any authors in this study.

\bibliographystyle{unsrt}
\bibliography{spinepose.bib}

\clearpage\appendix
\section{Online Methods}
\paragraph{Model Architecture.} The training pipeline of SpinePose uses a divide-and-conquer approach to detect keypoints in different regions on the X-ray images using three parallel models. ``Top-down" region-based models detect keypoints in the L1 and S1 regions, and a ``bottom-up" encoder-decoder model (R model) predicts the remaining keypoints (Figure 1) \cite{girshick2014rich, girshick2015fast, ren2015faster, he2017mask}. A region specific model is employed for the L1 and S1 regions due to a greater absolute number of keypoints to predict as well as fewer distinct anatomic features relative to structures such as the round femoral heads. Therefore, in each of these regions, the model first detects a bounding box and then subsequently predicts keypoints within each bound. Conversely, the R model uses a bottom-up approach to directly predict the remaining keypoints without detecting a region of interest (ROI), reducing computational complexity.

Each model uses a ResNet-50 feature pyramid network to extract features at multiple image scales. In each of the L1/S1 models, these features are used to learn a ROI using a region proposal network, and the features within the ROI are used to detect keypoints in each region (Figure 1). On the other hand, the R model does not rely on a bounding box detector, and instead directly predicts keypoints using learned features from the feature pyramid network. In each of the three models, a cross-entropy loss ($L_\text{Keypoint}$) over predicted binary keypoint maps is used to learn the position of the keypoints. Additional loss functions are used to supervise bounding box detection in the L1/S1 models, including a regression-based smooth L1 loss ($L_\text{Box}$) over predicted bounding box coordinates, and a binary cross-entropy loss ($L_\text{Class}$) on the predicted class of the object. 

Each of the three models was trained on a NVIDIA GeForce GTX 1080 GPU with a batch size of three. Each model was optimized using the Stochastic Gradient Descent optimizer, with a step-scheduled learning rate beginning at $10^{-3}$. Regularization methods included batch normalization, dropout, and image augmentation via rotation, brightness/contrast jitter, and translational shift. These augmentation strategies were used to improve model generalizability to the different image qualities and ASD severities often seen in clinical practice. The L1, S1, and R models were trained for 80, 40, and 100 epochs, respectively, and three different random train/validation splits were used to assess model reliability over different test image samples.

To extend the model's performance to lateral lumbosacral x-rays with only whole-spine x-rays in the training set, regularization methods consisted only of random cropping. Since the ultimate parameters of interest here were LL and SS, only the L1 and S1 models were trained. The training parameters were the same as those used during whole-spine predictions except for training time where the L1 and S1 models were trained for 80 and 60 epochs, respectively. 

\end{document}